\documentclass[10pt,journal, compsoc]{IEEEtran}
\ifCLASSOPTIONcompsoc
 \usepackage[nocompress]{cite}
\else
 \usepackage{cite}
\fi
\usepackage[switch]{lineno}

\usepackage{amsmath}
\usepackage{amsfonts}

\usepackage{etoolbox} 

\newcommand*\linenomathpatch[1]{%
  \cspreto{#1}{\linenomath}%
  \cspreto{#1*}{\linenomath}%
  \csappto{end#1}{\endlinenomath}%
  \csappto{end#1*}{\endlinenomath}%
}

\linenomathpatch{equation}
\linenomathpatch{align}

\usepackage{color}
\usepackage[normalem]{ulem}
\usepackage{subfigure}

\ifCLASSINFOpdf
\usepackage[pdftex]{graphicx}
\else
\fi

\begin{document}
%
\title{Short and Long Range Relation\\Based Spatio-Temporal Transformer for Micro-Expression Recognition}

%
%
%
%

\author{Liangfei~Zhang,
 Xiaopeng~Hong,~\IEEEmembership{Member},
 Ognjen~Arandjelovi\'c,~\IEEEmembership{Member}
 and~Guoying Zhao,~\IEEEmembership{Fellow}
\IEEEcompsocitemizethanks{\IEEEcompsocthanksitem L. Zhang and O. Arandjelovi\'c are with the School of Computer Science, University of St Andrews, UK.\protect\\
E-mail: lz36@st-andrews.ac.uk, oa7@st-andrews.ac.uk
\IEEEcompsocthanksitem X. Hong is with  Harbin Institute of Technology, P.R.China.
E-mail: hongxiaopeng@ieee.org
\IEEEcompsocthanksitem G. Zhao is with  University of Oulu, Finland. 
E-mail: guoying.zhao@oulu.fi}
}

\IEEEtitleabstractindextext{%
\begin{abstract}
Being spontaneous, micro-expressions are useful in the inference of a person's true emotions even if an attempt is made to conceal them. Due to their short duration and low intensity, the recognition of micro-expressions is a difficult task in affective computing. The early work based on handcrafted spatio-temporal features which showed some promise, has recently been superseded by different deep learning approaches which now compete for the state of the art performance. Nevertheless, the problem of capturing both local and global spatio-temporal patterns remains challenging. To this end, herein we propose a novel spatio-temporal transformer architecture -- to the best of our knowledge, the first purely transformer based approach (i.e.\ void of any convolutional network use) for micro-expression recognition. The architecture comprises a spatial encoder which learns spatial patterns, a temporal aggregator for temporal dimension analysis, and a classification head. A comprehensive evaluation on three widely used spontaneous micro-expression data sets, namely SMIC-HS, CASME II and SAMM, shows that the proposed approach consistently outperforms the state of the art, and is the first framework in the published literature on micro-expression recognition to achieve the unweighted F1-score greater than 0.9 on any of the aforementioned data sets.
\end{abstract}

}

\maketitle

\nolinenumbers 
\section{Introduction}\label{sec:introduction}
Facial expressions play an important role in interpersonal communication and their recognition is one of the most significant tasks in affective computing. Though there some disagreement on this remains, a notable number of psychologists believe that although due to different cultural environments individuals use different languages to communicate, the expression of their emotions is rather universal~\cite{Ekman1971}. Correctly recognizing facial expressions is important in general communication and can help understanding people's mental state and emotions.

When colloquially used, the term `facial expressions' refers to what are more precisely technically termed \emph{facial macro-expressions}~(MaEs). While crucial for human interaction, providing a universal and non-verbal means of articulating emotion~\cite{zhang2020quantification}, facial macro-expressions can be effected voluntarily which means that they can be used to deceive. In other words, a person's macro-expression may not accurately represent their truly felt emotion. However, whatever the conscious effort, felt emotions effect short-lasting contraction of facial muscles which are expressed \emph{involuntarily} under psychological inhibition. The resulting minute, sudden, and transient expressions are referred to as \emph{micro-expressions}~(MEs). After being first observed and recognized as a phenomenon of interest by Haggard and Isaacs~\cite{Gottschalk1966}, and then further elaborated on by a case study reported by Ekman and Friesen~\cite{Ekman1969}, MEs began to be researched more widely by psychologists, and in the last decade attracting interest within the field of computer vision~\cite{Zhang2021}. In contrast to MaEs, MEs are subtle. They are exhibited for 0.04s to 0.2s~\cite{Ekman1971}, and with lesser facial movement. These characteristics make MEs harder to be recognized than MaEs, whether manually (i.e.\ by humans) or automatically (i.e.\ by computers). 

The seminal work by Pfister, \emph{et al.} and the release of the database of micro-expression movie clips, namely SMIC-sub (Spontaneous Micro-expression Corpus)~\cite{Pfister2011}, effected a marked empowerment of computer scientists in the realm of micro-expression recognition (MER). The first generation of solutions built upon the well-established computer vision tradition and introduced a series of handcrafted features, such as Local Binary Pattern-Three Orthogonal Planes~(LBP-TOP)~\cite{Zhao2007}, 3~Dimensional Histograms of Oriented Gradients~(3DHOG)~\cite{Polikovsky2009}, Histograms of Image Gradient Orientation~(HIGO)~\cite{Li2018} and Histograms of Oriented Optical Flow~(HOOF)~\cite{Liu2016} and their variations. The next generation shifted focus towards Convolutional Neural Network~(CNN) based deep learning methods~\cite{patel2016selective,Khor2018,Li2019,Xia2019,Xia2020}. Early work by and large uses convolutional kernels to extract spatial information from micro-expression video frames. This kind of pixel level operators can be considered as capturing ``\emph{short-range}'', local spatial relationships. ``\emph{Long-range}'', global relationships between different spatial regions have also been proposed and studied, notably by means of Graph Convolutional Network~(GCN) based architectures~\cite{Lo2020,Buhari2020,Xie2020,Kumar2021,Lei2021}. The activations of Facial Action Units~(AUs) are generally used as nodes to build graphs. The relationships between different AU engagements are combined with image features to improve the discriminatory power in the context of MER. However, though these approaches consider global spatial relations so as to assist learning, they can only learn these after local features are extracted, i.e.\ they are unable to learn both kinds of relations jointly.

\begin{figure}[h!]
\centering
\includegraphics[width=\columnwidth]{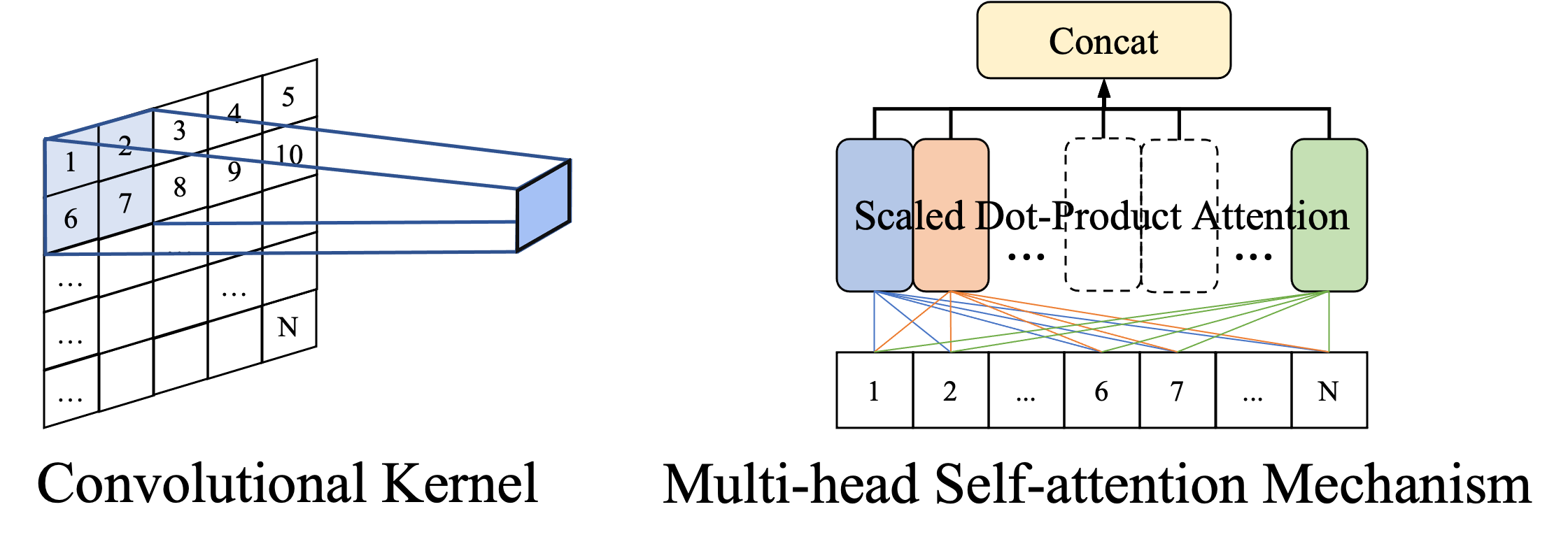}
\caption{Comparison of the different spatial feature extraction methods of CNN and transformer.} 
\label{fig:cnntranscom}
\end{figure}

In order to capture automatically both short- and long-range relations at the same time, we apply Multi-head Self-attention Mechanism~(MSM) instead of a Convolutional Kernel as the cornerstone of our deep learning MER architecture. As shown in Fig.~\ref{fig:cnntranscom}, the relations between block $1$ and $N$ will hardly ever be learnt by CNN but has been considered at the beginning of MSM. MSM based networks are called \emph{Transformer}. Short-range and long-range relationships between elements of a sequence can be learned in a parallelized manner because transformers utilize sequences in their entirety, as opposed to processing sequence elements sequentially like recurrent networks. Most recently, transformer networks came to the attention of the CV community. By dividing them into smaller constituent patches, two-dimensional images can be converted into one-dimensional sequences, translating the spatial relationships into the relationships between sequence elements~(image patches). In this way, transformer networks can be simply applied to vision problems and on various tasks they have outperformed CNNs~\cite{Khan2021}. Examples include segmentation~\cite{Ye2019}, image super-resolution~\cite{Yang2020}, image recognition~\cite{Dosovitskiy2020,Touvron2020}, video understanding~\cite{Sun2019,Girdhar2019} and object detection~\cite{Carion2020,Zhu2020}.  

Most MER research in the published literature is video based, as Ben et al.\ elaborated~\cite{ben2021}, though there is a small but notable body of work on single-frame analysis~\cite{Liong2018,Gan2019,Li2021}. This statistic reflects the consensus that for best performance both spatial and temporal information need be considered. In particular, absolute and relative facial motions are extracted and analysed through spatial and temporal features respectively. Most handcrafted methods in existence use the same kind of operator to detect spatial and temporal information from different dimensions by considering the frames as 3D data. The resulting spatio-temporal features with uniform format are used together to implement video based MER. In deep learning based methods, spatial features are mainly extracted by means of a convolutional neural network. Some concatenate spatial features extracted from each frame and others use recurrent neural networks to derive temporal information. To integrate various spatio-temporal relations, our design makes use of long-term temporal information in spatial data (i.e.\ each frame of video sample) prior to the spatial encoder, and a temporal aggregation block to fuse both short- and long-term temporal relationships afterwards.

In this work we show how a transformer based deep learning architecture can be applied to MER in a manner which outperforms the current state of the art. The main contributions of the present work are as follows:
\begin{figure}[h!]
\centering
\includegraphics[width=\columnwidth]{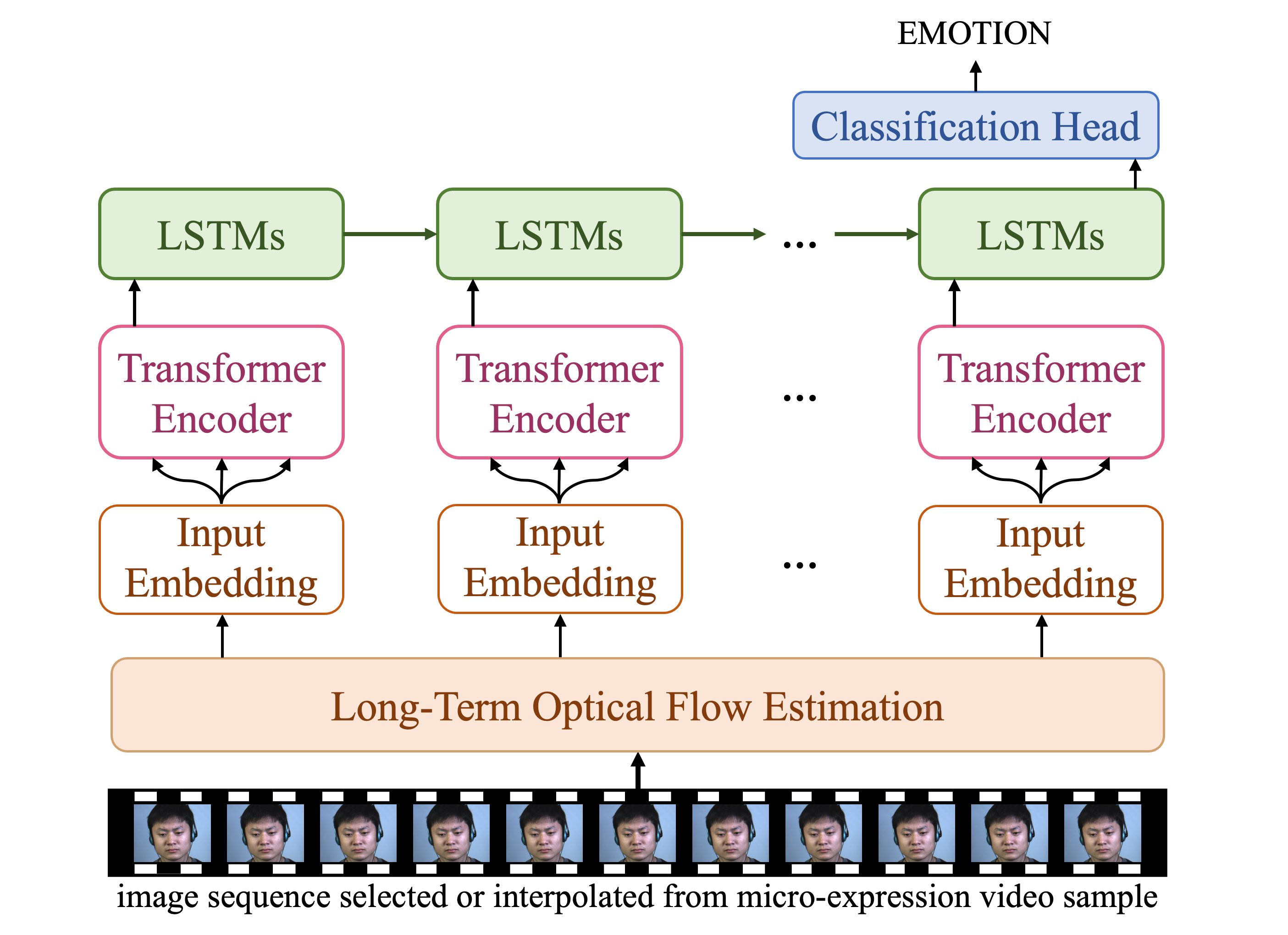}
\caption{The framework of proposed Short and Long range relation based Spatio-Temporal Transformer~(SLSTT).}
\label{fig:framework}
\end{figure}
\begin{enumerate}
 \item We propose a novel spatio-temporal deep learning transformer framework for video based micro-expression recognition, which we name \emph{Short and Long range relation based Spatio-Temporal Transformer~(SLSTT)}, the structure whereof is summarized in Fig.~\ref{fig:framework}. To the best of our knowledge, ours is the first deep learning MER work of this kind, in that it does not employ a CNN at any stage, but is rather entirely centred on a transformer architecture.

 \item We use matrices of long-term optical flow, computed in a novel way particularly suited for MER, instead of the original colour images as the input to our network. The feature ultimately arrived at combines long-term temporal information and short- and long-range spatial relations, and is derived by a transformer encoder block. 
 
 \item We design a temporal aggregation block to connect spatio-temporal features of spatial relations extracted from each frame by multiple transformer encoder layers and achieve video based MER. The empirical performance and analysis of mean and LSTM (long short-term memory) aggregators is presented too.
\end{enumerate}

We evaluate our approach on the three well known and popular ME databases, Spontaneous Micro-Expression Corpus (SMIC)~\cite{Li2013}, Chinese Academy of Sciences Micro-Expression II (CASME II)~\cite{Yan2014} and Spontaneous Actions and Micro-Movements (SAMM)~\cite{Davison2018}, in both Sole Database Evaluation~(SDE) and Composite Database Evaluation~(CDE) settings and achieve state of the art results.

\section{Related Work}

\subsection{Micro-expression Recognition}
Since the publication of the SMIC data set in 2013, the volume of research on automatic micro-expression recognition has been increasing steadily over the years. From the handcrafted computer vision methods in the early years to the deep learning approaches more recently, the main ideas of micro-expression feature extraction could be categorized as primarily pursuing either a spatial strategy or a temporal one.

\subsubsection{Spatial Features}
The fundamental challenge of computer vision is that of extracting semantic information from images or videos. Whatever the approach, the extraction of some kind of spatial features is central to addressing this challenge. Micro-expression recognition is no exception. In a manner similar to many gradient based features applied previously on generic computer vision tasks, Polikovsky et al.~\cite{Polikovsky2009} proposed the use of a gradient feature adapted to MER to describe local dynamics of the face. The magnitudes of local gradient projections in the $XY$ plane is used to construct histograms across different regions, which are used as spatial features. LBP quickly became the most popular operator for micro-expression analysis after Pfister et al.\ \cite{Pfister2011} first applied it to MER. This operator describes local appearance in an image. The key idea behind it is that the relative brightness of neighbouring pixels can be used to describe local appearance in a geometrically and photometrically robust manner. Its widespread use and favourable performance often make it the default baseline method when new data sets are published, or a new ME related task proposed. As for deep learning approaches, CNN model can be thought as a combination of two components: a feature extraction part and a classification part. The convolution and pooling layers perform spatial feature extraction.

Further to local appearance based features, numerous other strategies have been described for spatial feature extraction in micro-expression analysis. One of the simplest and commonest of these employs facial Region Of Interest (ROI) segmentation. Polikovsky et al.~\cite{Polikovsky2009} segmented each face sample into 12 regions according to the Facial Action Coding System (FACS)~\cite{Ekman2002}, each region corresponding to an independent facial muscle complex, and applied appearance normalization to individual regions. Others have modified or extended this strategy, e.g.\ employing different methods for segmentation or different salient regions -- 11~\cite{Zhang2021a}, 16~\cite{Wang2014b}, 36~\cite{Liu2016} instead of 12 of Polikovsky et al. Spatial feature operators are applied with each ROI rather the whole image, thus providing a more nuanced description of the face. In recent years, a more principled equivalent of this strategy (in that it is learnt, rather than predetermined by a human), can be found in the form of attention blocks applied within neural networks to improve their ability to learn spatial features. These blocks can generate weight masks for feature maps, helping a network pay greater attention to significant regions. Most recently, GCNs have also been used within deep learning frameworks as a means of capturing spatial information, often using AUs as correponding to graph nodes. For example, Lei et al.~\cite{Lei2021} segment node patches based on facial landmarks and fuse them with an AU GCN. Xie et al.~\cite{Xie2020} infer AU node features from the backbone features by global average pooling and use them to build an AU relation graph for GCN layers. These optimization measures use a priori knowledge (AUs in FACS) to enhance the extracted spatial features. Long-range spatial relationships are not directly learnt by such networks. 

\subsubsection{Temporal Features}
Since one of the most characteristic aspects of micro-expressions is their sudden occurrence, temporal features cannot be ignored. While some methods in the literature do use only the single, apex frame instead of all frames in each ME sample~\cite{Peng2019a,Gan2019,Liong2018,Li2021}, most employ all in the range between the onset frame and the offset, thus treating all temporal changes within this time period on the same footing. Some go further and employ temporal frame interpolation (as indeed we do herein) so as to increase the frame count~\cite{Li2018,Liu2016,Khor2018,Wang2014b,Pfister2011}.

A vast number of handcrafted feature based approaches treat raw video data as a 3D spatio-temporal volume, treating the temporal dimension as no different than the spatial ones. In other words, they  apply the same kind of operator used to extract spatial features on pseudo-images formed by a cut through the 3D volume comprising one spatial dimension and the temporal dimension. For example, in LBP-TOP, LBP operators are applied on $XT$ and $YT$ planes to extract temporal features, and their histogram across the three dimensions forms the final representation. 3DHOG similarly treats videos as spatio-temporal cuboids with no distinction made between the three dimensions, but arguably with even greater uniformity than LBP-TOP in that the descriptor itself is inherently 3D based. Similar in this regard are optical flow based features, which too inherently combine local spatial and temporal elements -- the use of optical strain~\cite{Liong2019}, flow orientation~\cite{Liu2016} or its magnitude~\cite{Liong2018} are all variations on this theme. 

As an alternative to the use of raw appearance imagery as input to a deep learning network, the use of pre-processed data in the form of optic flow matrices has been proposed by some authors~\cite{Xia2020,Liu2019,Kumar2021}. In this manner, proximal temporal information is exploited directly. On the other hand, the learning of longer range temporal patterns has been approached in a variety of ways by different authors. Some extract temporal patterns simply by treating video sequences as 3-dimensional matrices~\cite{Lo2020,Liong2019,Reddy2019}, rather than 2-dimensional ones which naturally capture single images. Others employ structures such as the recurrent neural network~(RNN) or the LSTM~\cite{Kim2016a,Khor2018}. In addition to the use of off-the-shelf recurrent deep learning strategies, recently there has been an emergence of methods which apply domain specific knowledge so as to make the learning particularly effective for micro-expression analysis~\cite{Xia2020}.

\subsection{Transformers in Computer Vision}
For approximately a decade now, convolutional neural networks have established themselves as the backbone of most deep learning algorithms in computer vision. However, convolution always operates on fixed size windows and is thus unable to extract distal relations. The idea of a transformer was first introduced in the context of NLP. It relies on a self-attention mechanism, learning the relationships between elements of a sequence. Transformers are able to capture `long-term’ dependence between sequence elements which is challenging for conventional recurrent models to encode. By dividing an image into sub-images and imposing a consistent ordering on them, a planar image can be converted into a sequence, so spatial dependencies can be learned in the same way as temporal features. For this reason, transformer based deep learning architectures have recently gained significant attention from the computer vision community and are starting to play an increasing role in a number of computer vision tasks.

A representative example in the context of object detection is the DEtection TRansformer (DETR)~\cite{Carion2020} framework which uses transformer blocks first, for regression and classification, but the visual features are still extracted by a CNN based backbone. The Image Generative Pre-Training (iGPT) approach of Chen et al.~\cite{Chen2021} attempts to exploit the strengths of transformers somewhat differently, pre-training BERT (Bidirectional Encoder Representations from Transformers)~\cite{Devlin2018}, originally proposed for language understanding, and thereafter fine tuning the network with a small classification head. iGPT uses pixels instead language tokens within BERT, but suffers from significant information loss effected by a necessary image resolution reduction. In the context of classification, the Vision Transformer (ViT) approach of Dosovitskiy et al.~\cite{Dosovitskiy2020} applies transformer encoding of image patches as a means of extracting visual features directly. It is the first pure vision transformer, and in its spirit and design, follows the original transformer~\cite{Vaswani2017} architecture faithfully. As such, it facilitates the application of scalable transformer architectures used in NLP effortlessly. 

Following these successes, transformers have been applied to a variety of computer vision tasks, including those in the realm of affective computing~\cite{Chen2020,Wang2021}. Notable examples include facial action unit detection~\cite{Jacob2021} and facial image-based macro-expression recognition~\cite{Ma2021}. However, none of the existing approaches to micro-expression recognition adequately make use of both the spatial and temporal information due to the design difficulties posed by the challenges we discussed in the previous sections. 

\section{Proposed Method}
In the present work we propose a method that takes advantage both of the physiological understanding of micro-expressions and their characteristics, as well as of the transformer framework. The approach overcomes many of the weaknesses of the existing MER methods in the literature as discussed in the previous section. Importantly, our method is able to extract and thus benefit both from proximal (i.e.\ short-range) and distal (i.e.\ long-range) spatio-temporal features. Each element of the proposed framework is laid out in detail next, corresponds to each sub-section.

\subsection{Long-term Optical Flow}
\label{sec:OF}
Optical flow describes the apparent motion of brightness patterns between frames, caused by the relative movement of the content of a scene and the camera used to image it~\cite{pham2014detection}.
If the camera is static, optical flow can be used to infer both the direction and the magnitude of an imaged object's movement from the change in the appearance of pixels between frames~\cite{arandjelovic2015cctv}.

\begin{figure}[h!]
\centering
\includegraphics[width=\columnwidth]{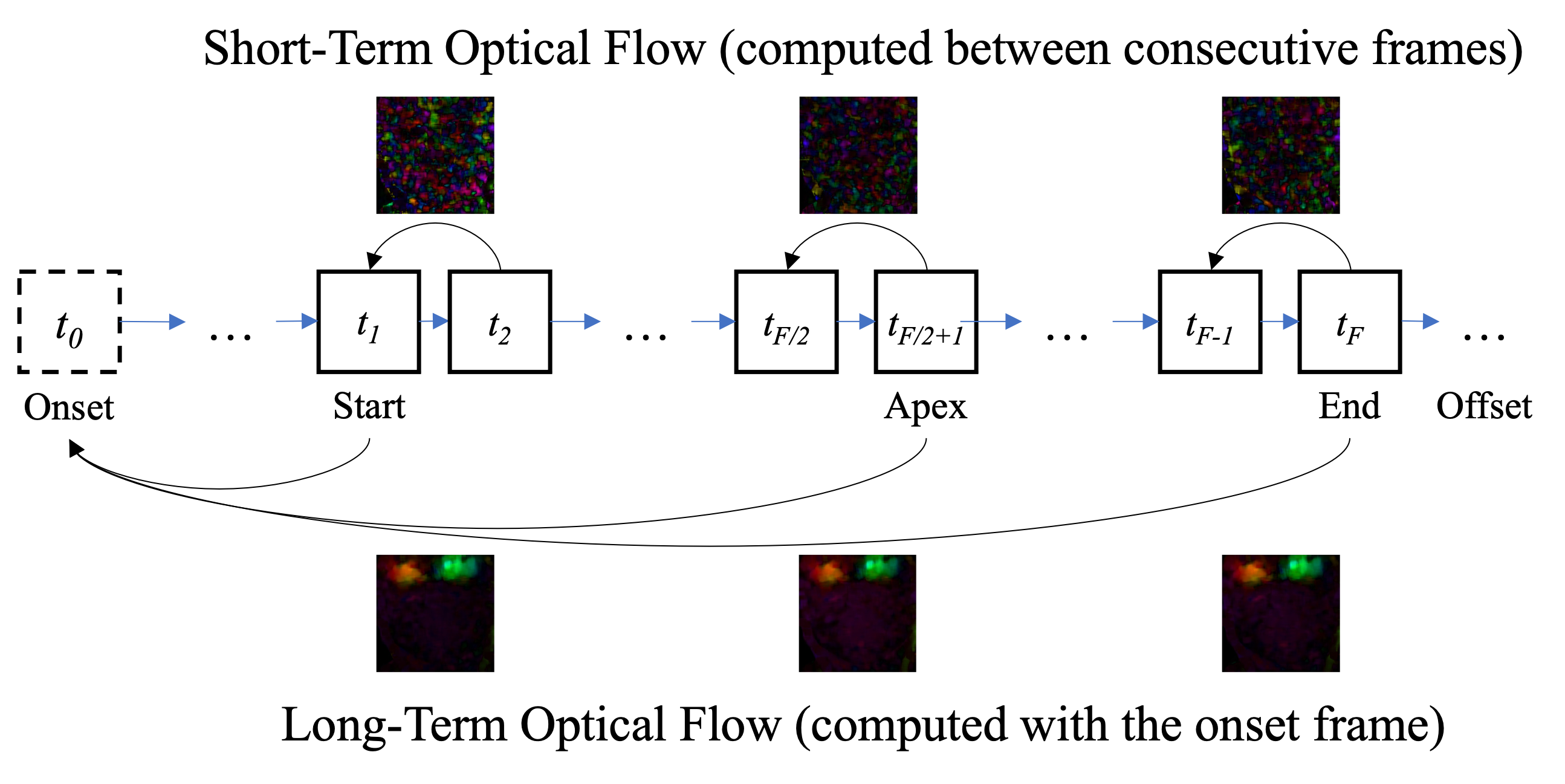}
\caption{Different computing mechanism between short- and long-term optical flow.}
\label{fig:OF}
\end{figure}

\begin{figure}[h!]
  \centering
  \subfigure[Onset frame, magnified region of interest] {\includegraphics[width=0.23\textwidth]{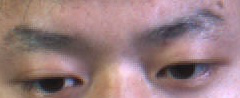}}
  \subfigure[Apex frame with superimposed long-term optical flow samples]{\includegraphics[width=0.23\textwidth]{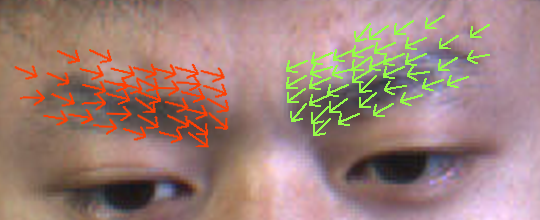}}
  \subfigure[Magnitude of long-term optical flow between onset \& apex frames] {\includegraphics[width=0.23\textwidth]{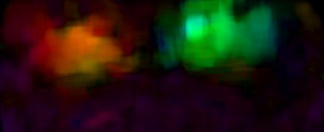}}
  \subfigure[Magnitude of short-term optical flow between apex frame and its previous one] {\includegraphics[width=0.23\textwidth]{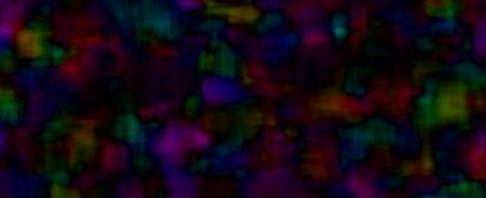}}
\caption{Illustration of optic flow computed between the onset and the apex frame, corresponding to the motion effected by the activation unit Brow Lowerer~(AU4).Compare with the one computed between consecutive frames.}
\label{fig:pixelmoving}
\end{figure}

Optical flow is inherently temporally local, i.e.\ save for practical considerations (numerical, efficiency, etc.) it is computed between consecutive frames of sequence. This introduces a problem when micro-expression videos are considered, created by the already noted limited motion exhibited during the expressions. Therefore, herein we propose to calculate optical flow between each sample frame and the onset frame instead of consecutive frames, see Fig.~\ref{fig:OF}. To see the reasons behind this choice, consider Fig.~\ref{fig:pixelmoving} which shows optical flow fields of consecutive frames starting with the micro-expression onset frame. It can be readily observed that the fields are rather similar up to the apex frame, which can be attributed to the aforementioned brevity of the expression, with a similar trend thereafter but in the opposite direction. In contrast, our, temporally non-local modified optical flow -- long-term optical flow in a manner of speaking -- exhibits a much more structured pattern, always being in the same direction, increasing in magnitude up to the apex frame and declining in magnitude thereafter. This results in much more stable and discriminative features associated with each micro-expression.

\begin{figure*}[ht]
\centering
\includegraphics[width=0.9\textwidth]{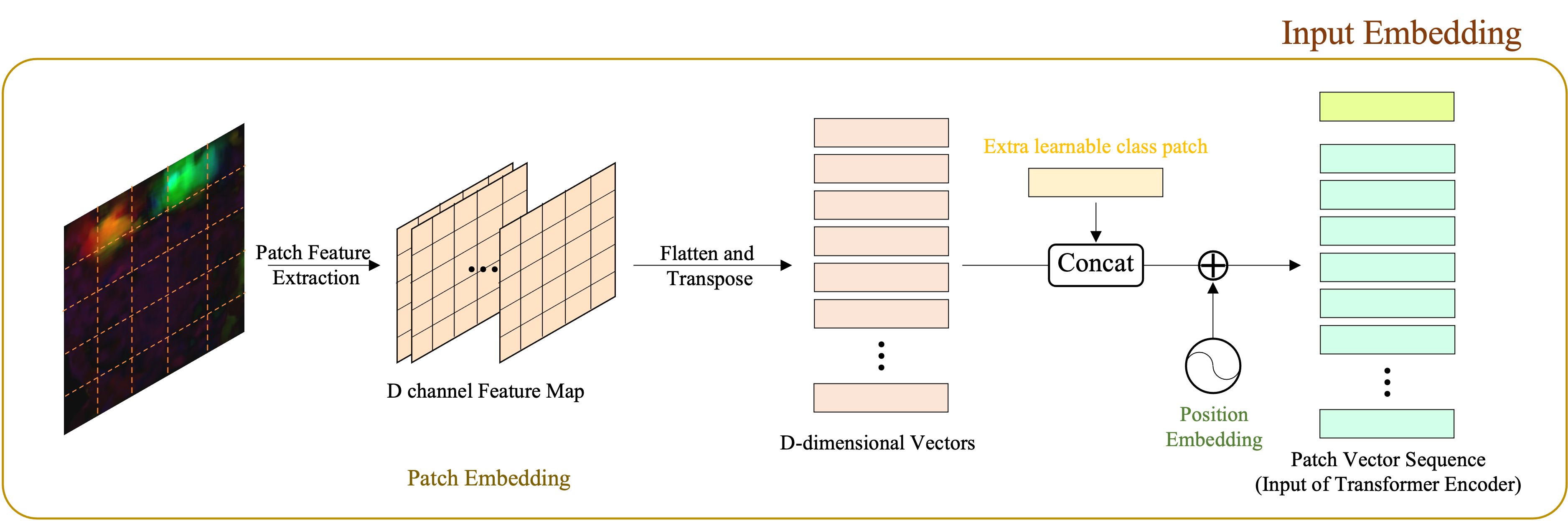}
\caption{Long-term optical flow fields are as inputs of the Input Embedding blocks. After short-range spatial feature extraction, patch and position embedding, the resulting sequence of vectors are fed to standard transformer encoder layers.}
\label{fig:embedding}
\end{figure*}

\subsection{Spatial Feature Extraction}
The key idea underlying the proposed method lies in the extraction of long-range spatial relations from each frame using a transformer encoder, with images as before being treated as sequences of constituent patches. More specifically, input frames are first represented as vector sequences with local spatial features of each image patch. The resulting sequences are then fed into the transformer encoder for long-term spatial feature extraction. 

\subsubsection{Input Embedding and Short-Range Spatial Relation Learning}
The standard transformer receives a 1D sequence as input. To handle 2D images, we represent each image as a sequence of rasterized 2D patches. Herein we do not use appearance images, that is the original video sequence frames, as input but rather the corresponding optical flow fields. An input embedding block is proposed as a means of representing input images as vector sequences for input to the transformer encoder. 

The general input embedding mechanism considers the image $X \in \mathbb{R}^{H \times W \times C}$ as a sequence of non-overlapping $P\times P$ pixel patches, where $H$, $W$, and $C$ are respectively the height, the width, and the channel count of the input. Different from the ``separate and flat'' linear patch embedding proposed by Dosovitskiy et al.~\cite{Dosovitskiy2020}, we first extract local spatial features in patch regions with a patch-wise fully connected layer. Patches of image $X$ are represented as $X_p \in \mathbb{R}^{N \times (P^2,C)}$. As shown in Fig.~\ref{fig:embedding}, we extract the short-range spatial features from image $X$ to feature map $X \in \mathbb{R}^{\frac{H}{P} \times \frac{W}{P} \times D}$, flatten and transpose them to $N$ $D$-dimensional vectors, where $N = \frac{HW}{P^2}$ the resulting number of patches in each image. $D$-dimensional vectors are passed through all transformer encoder layers. The specific values of parameters used in our experiments are stated in Section~\ref{sec:exp}. 

After that, a learnable $D$-dimensional vector is concatenated with the sequence, as the class token ($Z_0[0] = x_{class}$), whose state as the output of the transformer encoder ($Z_{L_T}[0]$). The effective input sequence length for the transformer encoder is thus $N+1$. Then a position embedding is added to each vector in the sequence. The whole input embedding procedure can be described as follows: 
\begin{align} 
\label{eq:LP} 
Z_0 = [X_{class}; X_p^1E;X_p^2E;\dots;X_p^NE]+E_{pos},\nonumber\\
E \in \mathbb{R}^{(P^2,C)\times D}, E_{pos} \in \mathbb{R}^{(N+1)\times D}, 
\end{align} 
where $Z_0 \in \mathbb{R}^{(N \times D)}$ is the input of the transformer encoder.

\begin{figure}[h!]
\centering
\includegraphics[width=0.8\columnwidth]{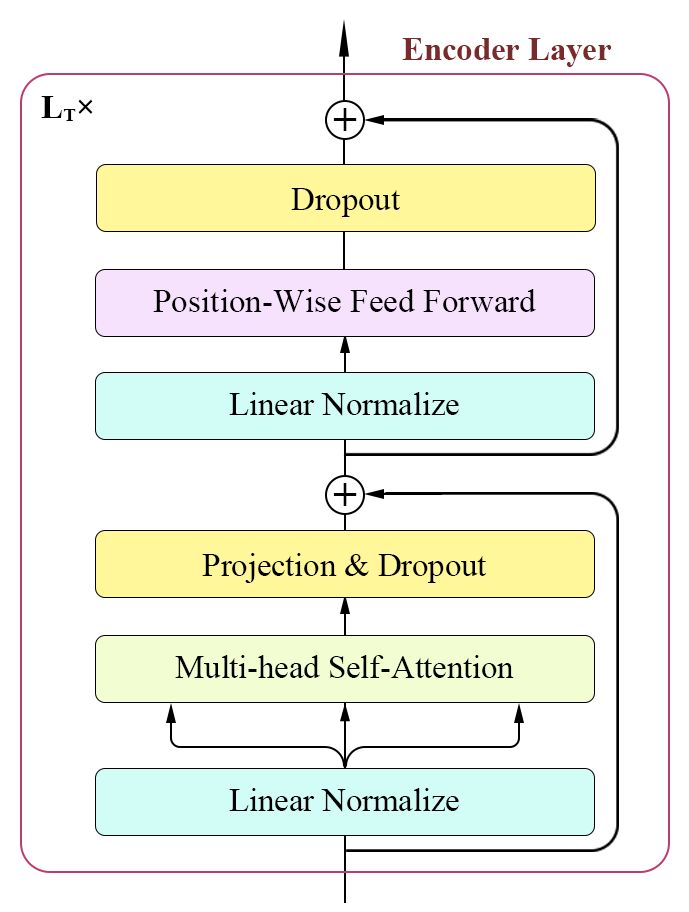}
\caption{Detailed structure of a Transformer Encoder layer. The output of frame $t$ processed by spatial encoder is $Z_{L_T}^t$.}
\label{fig:encoderlayer}
\end{figure}

\begin{figure*}[ht]
\centering
\includegraphics[width=\textwidth]{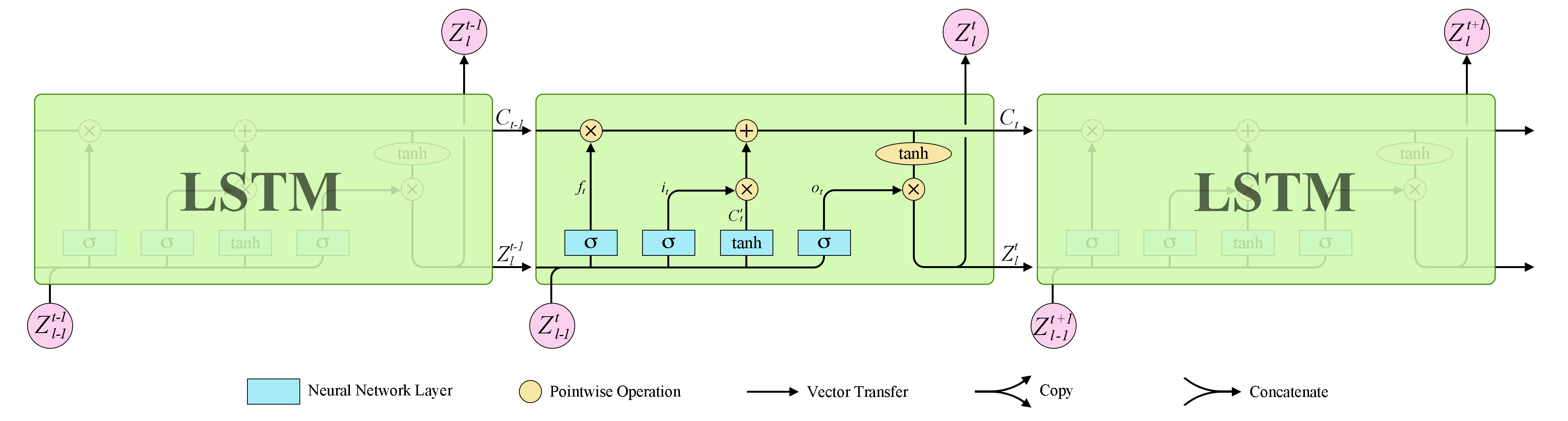}
\caption{The repeating module in an LSTM aggregator layer.}
\label{fig:LSTM}
\end{figure*}

\subsubsection{Long-Range Spatial Relation Learning by Transformer Encoder}
After short-range spatial relation are extracted from the input long-term optical flow fields of each frame and embedded as vectors, they are passed to a transformer encoder for further long-range spatial feature extraction. Our encoder contains $L_T$ transformer layers; herein we use $L_T=12$, adopting this value from the ViT-Base model of Dosovitskiy et al.~\cite{Dosovitskiy2020} (the pre-trained encoder we use in experiments).Each layer involves two blocks, a Multi-head Self-attention Mechanism~(MSM) and a Position-Wise fully connected Feed-Forward network~(PWFF), as shown in Fig.~\ref{fig:encoderlayer}. Layer Normalisation~(LN) is applied before each block and residual connections after each block~\cite{wang2019, Baevski2019}. The output of the transformer layer can be written as follows: 
\begin{align} 
Z_l' &= MSM(LN(Z_{l-1}))+Z_{l-1}, l = 1 \dots L_T, \\ 
Z_l &= PWFF(LN(Z_l'))+Z_l', l = 1 \dots L_T, 
\end{align} 
where $Z_l$ is the output of layer $l$. The PWFF block contains two layers with the Gaussian Error Linear Unit (GELU) non-linear activation function. The feature embedding dimension thereby first increases from $D$ to $4D$ and then reduces back to $D$, which equals 768 in our experiments. 

Multi-head attention allows the model to focus simultaneously on information content from different parts of the sequences, so both long-range and short-range spatial relations can be learnt. An attention function is mapping a query and a set of key-value pairs to the output, a weighted sum of the values. The weights are computed using a compatibility function of the queries with the corresponding keys, and they are all vectors. The self-attention function is computed on a set of queries simultaneously. The queries, keys and values can be grouped together and represented as matrices $Q$, $K$ and $V$, so the computation of the matrix of outputs can be written as: 
\begin{align} 
Q &= Z_{l-1}W_Q,\\
K &= Z_{l-1}W_K,\\
V &= Z_{l-1}W_V,
\end{align} 
\begin{equation}
SA(Z_{l}) = \text{softmax}\left(\frac{QK^T}{\sqrt{D}}\right)V,
\end{equation}
where $W_Q, W_K, W_V \in \mathbb{R}^{D \times D_m}$ are learnable matrices and SA is the self-attention module.
MSM can be seen as a type of self-attention with $M$ heads in parallel operation and a projection of their concatenated outputs:
\begin{equation} 
MSM(Z_{l}) = \text{Concat}( \left\{SA_h(Z_{l}), \forall h \in \left[1..M\right] \right\} ) W_O, 
\end{equation} 
where $W_O \in \mathbb{R}^{M\cdot D_m \times D}$ is a re-projection matrix. $D_m$ is typically set to $\frac{D}{M}$, so as to keep the number of parameters constant with changing $M$.

\subsection{Temporal Aggregation}
After extracting both local and global spatial features associated with each frame using a transformer encoder, we introduce an aggregation block to extract temporal features before performing the ultimate classification. The aggregation function ensures that our transformer model can be trained and applied to the spatial feature sets of each frame, subsequently processing the temporal relations between frames in each sample. Since facial movement during micro-expressions is almost imperceptible, all frames from a single video sample are rather similar one to another. Nevertheless, it is still possible to identify reliably a number of salient frames, such as the apex frame, that play a particularly important role in the analysis of a micro-expression. Therefore, we propose an LSTM architecture for temporal aggregation.

\emph{Long Short-Term Memory}~(LSTM)~\cite{Hochreiter1997} is a type of recurrent neural network with feedback connections, which overcomes two well-known problems associated with RNNs: the vanishing gradient problem, and the sensitivity to the variation of the temporal gap length between salient events in a processed sequence. The elements of the input are the sets of outputs from the transformer encoder for each frame. The inputs are not concatenated, and the input sequence length is thus dependent on the number of frames in each ME video sample.

We used three LSTM layers in the aggregation block. The computation details of each layer are: 
\begin{align} 
  t&=1 \dots F, l=L_T+1 \dots L_A, \nonumber\\
  f_t &= \sigma (W_f \cdot [Z_{l}^{t-1},Z_{l-1}^{t}]+b_f), \\ 
  i_t &= \sigma (W_i \cdot [Z_{l}^{t-1},Z_{l-1}^{t}]+b_i), \\ 
  o_t &= \sigma (W_o \cdot   [Z_{l}^{t-1},Z_{l-1}^{t}]+b_o), 
\end{align} 
\begin{align}
  C_t' &= tanh(W_C \cdot [Z_{l}^{t-1},Z_{l-1}^{t}] +b_C),\\ 
  C_t &= f_t \times C_{t-1} + i_t \times C_t',
\end{align} 
\begin{align}
  Z_l^t &= o_t \times tanh(C_t),
\end{align} 
where $F$ is the number of chosen frames in each video sample, $L_A$ is the total number of layers in both the transformer encoder and the LSTM aggregator. $Z_l^t$ denotes the outputs of the layer $l$ after $t$ frames have been processed. After all frames are processed in this manner, the result is a single feature set describing the entire micro-expression video sample. Finally, these features are fed into an MLP which is used for the ultimate MER classification. The details of how previous output join the latter training are presented in Fig.~\ref{fig:LSTM}. We also design a comparative experiment to demonstrate the effectiveness of the LSTM aggregator, the details of which are described in the Section~\ref{sec:tempExp}.

\subsection{Network Optimization}
Following the aggregation block, our network contains two fully connected layers which facilitate the final classification achieved using the SoftMax activation function. Cross Entropy loss is used as the objective function for training: 
\begin{equation}
  L = \frac{1}{N} \sum_i L_i = - \frac{1}{N} \sum_i\sum_{c=1}^{C} y_{ic} \log(p_{ic}),
\end{equation}
where $N$ is the number of the ME video samples and $C$ the number of emotion classes. The value of $y_{ic}$ is 1 when the true class of sample $i$ is equal to $c$ and 0 otherwise. Similarly, $p_{ic}$ is the predicted probability that sample $i$ belongs to class $c$.

When using gradient descent to optimize the objective function during network training, as the parameter set gets closer to its optimum, the learning rate should be reduced. Herein we achieve this using cosine annealing~\cite{Loshchilov2017}, i.e.\ using the the cosine function to modulate the learning rate which initially decreases slowly, and then rather rapidly before stabilizing again. This learning rate adjustment is particularly important in the context of the problem at hand, considering that the number of available micro-expression video samples is not large even in the largest corpora, readily learning to overfitting if due care is not taken.

\section{Experiments and Evaluation}
\label{sec:exp}
In this section we describe the empirical experiments used to evaluate the proposed method. We begin with a description of the data sets used, follow up with details on the data pre-processing performed, relevant implementation details, and evaluation metrics, and conclude with a report of the results and a discussion of the findings.

\subsection{Databases}
Following the best practices in the field, for our evaluation we adopt the use of three large data sets, namely the Spontaneous Micro-Expression Corpus~(SMIC)~\cite{Li2013}, the Chinese Academy of Sciences Micro-Expression II data set (CASME II)~\cite{Yan2014}, and  the Spontaneous  Actions  and  Micro-Movement database (SAMM)~\cite{Davison2018}, thus ensuring sufficient diversity of data, evaluation scale, and ready and fair comparison with other methods in the literature. All video samples in these databases capture spontaneously exhibited, rather than acted micro-expressions (see Zhang and Arandjelovi\'c~\cite{Zhang2021} for discussion), which is important for establishing the real-world applicability of findings.

\subsubsection{SMIC}
The Spontaneous Micro-Expression Corpus~(SMIC) is the earliest published spontaneous micro-expression database~\cite{Li2013}. It comprises three distinct parts captured by cameras of different types, namely a conventional visual camera~(VIS), a near-infrared camera~(NIR) and a high-speed camera~(HS). These subsets are designed to study micro-expression analysis tasks in various application scenarios. To achieve uniformity with the other two corpora, namely CASME-II and SAMM which are described next, which only contain high-speed camera videos, it is the HS subset from SMIC that we make use of herein. The SMIC-HS contains 164 video sequences (samples) from 16 subjects of 3 ethnicities. Using two human labellers, these videos are categorized as corresponding to either negative~(70), positive~(51), or surprised~(43) expression, and both raw and cropped frames are provided. 

\subsubsection{CASME II}
The Chinese Academy of Sciences Micro-Expression II~(CASME II) data set contains 247 micro-expression video samples from 26 Chinese participants. The full videos have the resolution of $640 \times 480$ pixels. Cropped facial frames in $280 \times 340$ pixel resolution (higher than both CASME and SMIC-HS), extracted using the same face registration and alignment method as for SMIC, are also provided. The micro-expression samples in CASME II are labelled by 2 coders to 5 classes, namely Happiness~(33), Disgust~(60), Surprise~(25), Repression~(27), and Others~(102). 

\subsubsection{SAMM}
The Spontaneous Actions and Micro-Movement (SAMM) database is the newest MER corpus. The 159 micro-expression short videos in the corpus were collected using 32 participants of 13 ethnicities, with an even gender distribution (16 male and 16 female), at 200~fps and the resolution of $2040 \times 1088$ pixels, with the face region size being approximately $400 \times 400$ pixels. The samples are assigned to one of 8 emotion classes, namely Anger~(57), Happiness~(26), Other~(26), Surprise~(15), Contempt~(12), Disgust~(9), Fear~(8) and Sadness~(6). 

\subsection{Data Pre-Processing}
\subsubsection{Face Cropping}
As noted in the previous section, cropped face images are explicitly provided in both SMIC-HS and CASME II data sets, with the same registration method used in both; no cropped faces are provided as part of SAMM. In order to maintain data consistency across different databases, in our experiments we employ a different face extraction approach. In particular, we utilize the Ensemble of Regression Trees~(ERT)~\cite{Kazemi2014} algorithm implemented in DLib~\cite{King2009} to localize salient facial loci (68 of them) in a uniform manner regardless of which data set a specific video sample came from.  

In the case of SMIC-HS and CASME II videos, the original authors' face extraction process consists of facial landmarks detection in the first frame of a micro-expression clip and then the detected face being registered to the model face using a LWM transformation. Motivated by the short duration of MEs, the faces in all remaining frames of the video sample are registered using the same matrix.

However, in this paper we employ an alternative strategy. The primary reason lies in the need for sufficient and representative data diversity, which is particularly important in deep learning. In particular, the original face extraction method just described, often results in the close resemblance of samples which increases the risk of model overfitting. Therefore, herein we instead simply use a non-reflective 2D Euclidean transformation, i.e.\ one comprising only rotation and translation. By doing so, at the same time we ensure the correct alignment of salient facial points and maintain information containing facial contour variability.   

\begin{figure}[h!]
    \centering
    \includegraphics[width=0.7\columnwidth]{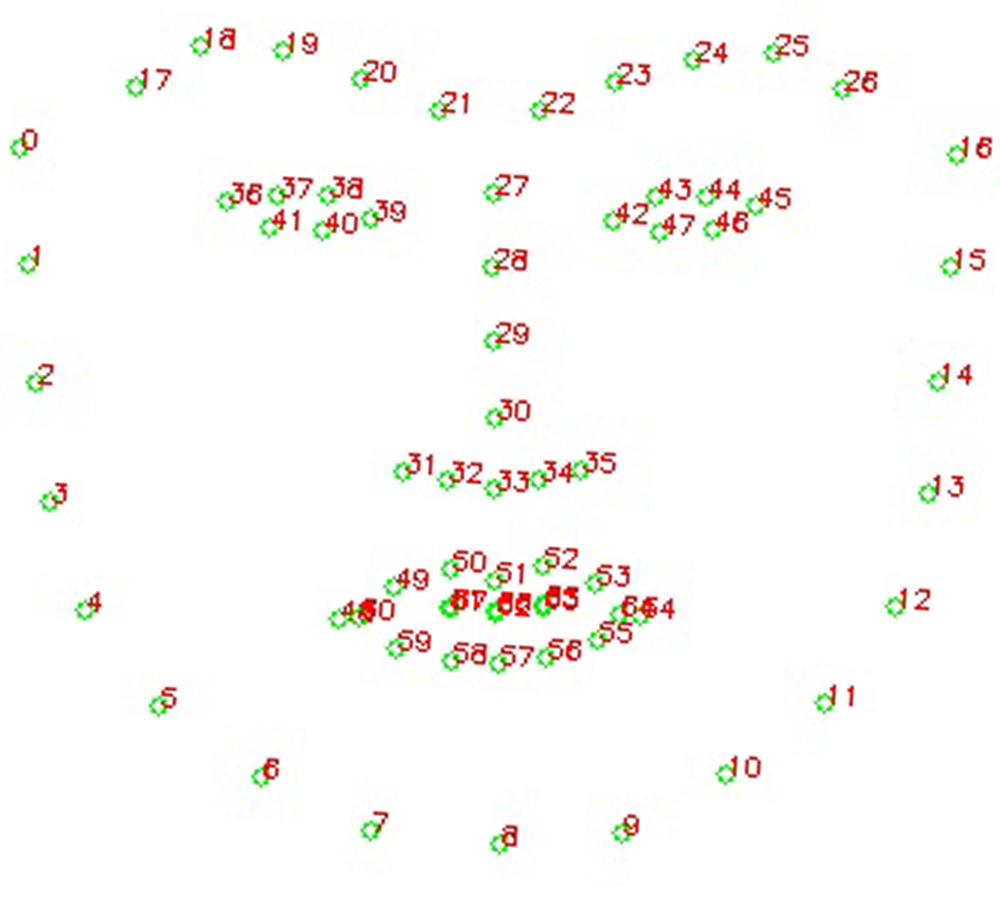}
    \caption{The 68 facial landmarks used by our method, shown for the onset (green) and the apex frame (red).}
    \label{fig:landmark}
\end{figure}

Furthermore, unlike the authors of SMIC-HS and CASME II, we do not perform facial landmark detection in the first frame of a micro-expression sample, but rather in the apex, thereby increasing the registration accuracy of the most informative parts of the video. As shown in Fig.~\ref{fig:landmark}, points 27–30 can be used to determine the centre line of the nose that can be considered as the vertical symmetry line of the entire face area. Point~30 is set as the centre point, and the square size $s$ (in pixels) is computed by adding the vertical distance from the centre point of the eyebrows~(19) to the lowest point of the chin~(8), $y_{apex[8]}-y_{apex[19]}$, to the height of chin, $y_{apex[8]}-y_{apex[57]}$, so that nearly the entire face is included in the cropped image:
\begin{equation}
s = (y_{apex[8]}-y_{apex[19]})+(y_{apex[8]}-y_{apex[57]}).
\end{equation}

\subsubsection{Temporal Interpolation}
Considering the short duration of micro-expressions, even when samples are acquired using high-speed cameras, in some instances only a small number (cc.\ 10) of frames is available. In an attempt to extract accurate temporal information, we also apply frame interpolation from raw videos, effectively synthetically augmenting data. In previous work, the Temporal Interpolation Model (TIM) relies on a path graph to characterize the structure of a sequence of frames, popularly used in several handcrafted feature based methods~\cite{Li2019,Wang2018a,Li2018}, whereas Liu et al.~\cite{Liu2016} use simple linear interpolation. Herein we propose a novel approach to interpolation so that its result is smoother in terms of optical flow, it being the nexus of our entire MER methodology. Most existing optical flow based methods produce artifacts on motion boundaries by estimating bidirectional optical flows, scaling and reversing them to approximate intermediate flows. We adopt the Real-time Intermediate Flow Estimation~(RIFE) method~\cite{Zhewei2020}, which uses an end-to-end trainable neural network, IFNet, which speedily and directly estimates the intermediate flows. 

Original RIFE interpolates one frame between two given consecutive frames, so we apply it recursively to interpolate multiple intermediate frames. Specifically, given any two consecutive input frames $I_0, I_1$, we apply RIFE once to get intermediate frame $\hat{I}_{0.5}$ at $t = 0.5$. We then apply RIFE to interpolate between $I_0$ and $\hat{I}_{0.5}$ to get $\hat{I}_{0.25}$, and so on. In our experiment, we prioritize interpolation in the temporal vicinity of the apex frame. The interpolated queue can be expressed as $\left\{\hat{I}_{a-0.5},\hat{I}_{a+0.5},\hat{I}_{a-1.5},\hat{I}_{a+1.5},\ldots,\hat{I}_{o+0.5}\;or\;\hat{I}_{f-0.5}\right\}$, where $a$, $o$ and $f$ are frame indices of the apex, onset, and offset frames respectively. Recall that the apex frames are specified explicitly in CASME II and SAMM, and for SMIC-HS we choose the middle frame of each sample video as the apex. If the number of interpolation frames is lower than the reference count (the average number of frames in this period across the database), we use the same method on the updated frame sequence iteratively to generate further intermediate frames.

\subsection{Experimental Settings}
\subsubsection{Implementation Details}
In the spatial feature extraction procedure, we employed base ViT blocks, with 12 Encoder layers, hidden size of 768, MLP size of 3072, and 12 heads. For initialization, we use the official ViT-B/16 model~\cite{Dosovitskiy2020} pre-trained on ImageNet~\cite{Deng2009}. We resize our input images to $384 \times 384$ pixels and split each image into patches with $16 \times 16$ pixels, so that the number of patches is $24 \times 24$. $768$-dimensional vectors are passed though all transformer encoder layers. For temporal aggregation, we select 11 frames (apex, and five preceding and succeeding it) per sample as inputs for the mean aggregator and LSTM aggregator. We have tried other options with different number of frame, but it didn't work any better. We only use long-term optical flow in experiments, as motivated by the arguments discussed in Section~\ref{sec:OF}. For learning parameters, the initial learning rate and weight decay are set to be 1e-3 and 1e-4, respectively. The momentum for Stochastic Gradient Decent~(SGD) is set to 0.9, with the batch size 4 for all experiments. All the experiments were conducted with PyTorch. 

\subsubsection{Mean Versus LSTM Aggregator}
\label{sec:tempExp}
We compare our LSTM aggregator with an alternative which uses the simple mean operator for temporal aggregation. After each frame is processed by spatial encoder, the corresponding output is used in the computation by the mean aggregation layer (layer $L_T+1$):
\begin{align} 
  Z_{L_T+1}^t = \frac{t-1}{t}Z_{L_T+1}^{t-1}+\frac{1}{t}Z_{L_T}^t, t=1 \dots F,
\end{align} 
In a manner similar to that described previously in the context of the LSTM Aggregator, outputs of each frame from our transformer encoder are taken as inputs to the temporal feature extraction module. Compared to the mean operator, LSTM has the advantage of larger expressive capability, resulting in different extracted relationships between different frames. Within the specific context of our work, this means that its ability to distinguish between emotions is also different, with LSTM expected to perform better.

\subsubsection{Evaluation Metrics}
Following previous work and the Micro-Expressions Grand Challenges (MEGCs), we conducted experiments on SMIC-HS, CASME~II, and SAMM, evaluating the classification performance using the corresponding original emotion classes, as well as the composite corpus formed using all three data sets and relabelled using three classes as proposed in MEGC~2019~\cite{See2019}. All results are reported using LOSO cross-validation. Evaluation is repeated multiple times by holding out test samples of each subject group while the remaining samples are used for training. In this way we best mimic real-world situations and in particular assess the robustness to variability in ethnicity, gender, emotional sensitivity, etc.

\paragraph{Sole Database Evaluation (SDE)}
In the first part of our empirical evaluation, experiments are conducted on three databases individually, using the corresponding original emotion labels, excepting the very rare (and thus underrepresented) classes in CASME~II and SAMM. SMIC-HS uses 3 class labels whereas the other two sets both use 5. We use \emph{accuracy} and \emph{macro F1-score} to assess the recognition performance.

\paragraph{Composite Database Evaluation~(CDE)}
In the second part of our empirical evaluation, experiments are conducted on the composite database with 3 emotion classes~(negative, positive, and surprise). The composite database, that is the database obtained by merging SMIC, CASME~II, and SAMM contains the total of 68 subjects, 16 from SMIC, 24 from CASME~II and 28 from SAMM. LOSO cross-validation is applied on each database separately and together on the composite database. \emph{Unweighted F1-score~(UF1)}, also known as the \emph{macro F1-score} and \emph{Unweighted Average Recall~(UAR)} are used to assess performance:
\begin{align}
  UF1 = \textit{macro F1-score},\\
  UAR = \frac{\sum_{c=1}^C \frac{\sum_{i=1}^S TP_{i,c}}{N_c}}{C},
\end{align}
where $N_c$ is the total number of samples of class $c$ across all subjects.

\begin{table*}[htbp]
  \def\arraystretch{1.25}
  \centering
  \caption{SDE results comparison with LOSO on SMIC-HS (3 classes), CASME II (5 classes) and SAMM (5 classes). Best performances are shown in bold, second best by square brackets enclosure. (* Reported by Huang et al.~\cite{Huang2015a}, ** Reported by Khor et al.~\cite{Khor2019})}
  
  \begin{tabular}{|l|c|c|c|c|c|c|}
    \hline
     & \multicolumn{2}{c|}{ \textbf{SMIC-HS}} & \multicolumn{2}{c|}{ \textbf{CASME II}} & \multicolumn{2}{c|}{ \textbf{SAMM}}\\
     \hline
     & \textbf{Acc(\%)} & \textbf{F1} & \textbf{Acc(\%)} & \textbf{F1} & \textbf{Acc(\%)} & \textbf{F1} \\
     \hline
     \textbf{Handcrafted} & & & & & & \\
     LBP-TOP* & 53.66 & 0.538 & 46.46 & 0.424 & – & – \\
     LBP-SIP* & 44.51 & 0.449 & 46.56 & 0.448 & – & – \\
     STLBP-IP~\cite{Huang2015} (2015) & 57.93 & – & 59.51 & – & – & – \\
     STCLQP~\cite{Huang2015a} (2015) & 64.02 & 0.638 & 58.39 & 0.584 & – & – \\
     Hierarchical STLBP-IP~\cite{Zong2018} (2018) & 60.37 & 0.613 & – & – & – & – \\
     HIGO+Mag~\cite{Li2018} (2018) & 68.29 & – & 67.21 & – & – & – \\
     \hline
     \textbf{Deep Learning} & & & & & & \\
     AlexNet** & 59.76 & 0.601 & 62.96 & 0.668 & 52.94 & 0.426 \\
     DSSN~\cite{Khor2019} (2019) & 63.41 & 0.646 & 70.78 & 0.730 & 57.35 & 0.464 \\
     AU-GACN~\cite{Xie2020} (2020) & – & – & 49.20 & 0.273 & 48.90 & 0.310 \\
     MER-GCN~\cite{Lo2020} (2020) & – & – & 42.71 & – & – & – \\
     Micro-attention~\cite{Wang2020} (2020) & 49.40 & 0.496 & 65.90 & 0.539 & 48.50
     & 0.402 \\
     Dynamic~\cite{Sun2020} (2020) & \textbf{76.06} & 0.710 & 72.61 & 0.670 & – & – \\
     GEME~\cite{Nie2021} (2021) & 64.63 & 0.616 & {[}75.20{]} & {[}0.735{]} & 55.88 & 0.454 \\
\hline
\textbf{SLSTT-Mean~(Ours)} & 73.17 & {[}0.719{]} & 73.79 & 0.723 & {[}66.42{]} & {[}0.547{]} \\
\textbf{SLSTT-LSTM~(Ours)} & {[}75.00{]} & \textbf{0.740} & \textbf{75.81} & \textbf{0.753} & \textbf{72.39} & \textbf{0.640} \\
  \hline
  \end{tabular}
  \label{tab:SDE}
\end{table*}

\begin{table*}[htbp]
\def\arraystretch{1.25}
  \centering
  \caption{CDE results comparison with LOSO on SMIC-HS, CASME II, SAMM and composite database (3 classes). Best performances are shown in bold, second best by square brackets enclosure. (*Reported by See et al.~\cite{See2019}, **Reported by Xia et al.~\cite{Xia2020a})}
  \begin{tabular}{|l|c|c|c|c|c|c|c|c|}
    \hline
     & \multicolumn{2}{c|}{\textbf{Composite}} & \multicolumn{2}{c|}{\textbf{SMIC-HS}} & \multicolumn{2}{c|}{\textbf{CASME II}} & \multicolumn{2}{c|}{\textbf{SAMM}} \\
     \hline
      & \textbf{UF1} & \textbf{UAR} & \textbf{UF1} & \textbf{UAR} & \textbf{UF1} & \textbf{UAR} & \textbf{UF1} & \textbf{UAR} \\
     \hline
     \textbf{Handcrafted} & & & & & & & & \\
    LBP-TOP* & 0.588 & 0.579 & 0.200 & 0.528 & 0.703 & 0.743 & 0.395 & 0.410 \\
    Bi-WOOF* & 0.630 & 0.623 & 0.573 & 0.583 & 0.781 & 0.803 & 0.521 & 0.514 \\
    \hline
    \textbf{Deep learning} & & & & & & & & \\
    ResNet18** & 0.589 & 0.563 & 0.461 & 0.433 & 0.625 & 0.614 & 0.476 & 0.436 \\
    DenseNet121** & 0.425 & 0.341 & 0.460 & 0.333 & 0.291 & 0.352 & 0.565 & 0.337 \\
    Inception V3** & 0.516 & 0.504 & 0.411 & 0.401 & 0.589 & 0.562 & 0.414 & 0.404 \\
    WideResNet28-2** & 0.505 & 0.513 & 0.410 & 0.401 & 0.559 & 0.569 & 0.410 & 0.404 \\
    OFF-ApexNet*~\cite{Gan2019} (2019) & 0.720 & 0.710 & 0.682 & 0.670 & 0.876 & 0.868 & 0.541 & 0.539 \\
    CapsuleNet~\cite{VanQuang2019} (2019) & 0.652 & 0.651 & 0.582 & 0.588 & 0.707 & 0.701 & 0.621 & 0.599 \\
    Dual-Inception~\cite{Zhou2019} (2019) & 0.732 & 0.728 & 0.665 & 0.673 & 0.862 & 0.856 & 0.587 & 0.566 \\
    STSTNet~\cite{Liong2019} (2019) & 0.735 & 0.761 & 0.680 & 0.701 & 0.838 & 0.869 & 0.659 & 0.681 \\
    EMR~\cite{Liu2019} (2019) & 0.789 & 0.782 & \textbf{0.746} & \textbf{0.753} & 0.829 & 0.821 & \textbf{0.775} & {[}0.715{]} \\
    ATNet~\cite{Peng2019a} (2019) & 0.631 & 0.613 & 0.553 & 0.543 & 0.798 & 0.775 & 0.496 & 0.482 \\
    RCN~\cite{Xia2020a} (2020) & 0.705 & 0.716 & 0.598 & 0.599 & 0.809 & 0.856 & 0.677 & 0.698 \\
    AUGCN+AUFsuion~\cite{Lei2021} (2021) & {[}0.791{]} & \textbf{0.793} & 0.719 & {[}0.722{]} & {[}0.880{]} & {[}0.871{]} & {[}0.775{]} & \textbf{0.789}\\
  \hline
  \textbf{SLSTT-Mean~(Ours)} & 0.788 & 0.767 & 0.719 & 0.699 & 0.844 & 0.830 & 0.625 & 0.566 \\
  \textbf{SLSTT-LSTM~(Ours)} & \textbf{0.816} & {[}0.790{]} & {[}0.740{]} & 0.720 & \textbf{0.901} & \textbf{0.885} & 0.715 & 0.643 \\
  \hline
  \end{tabular}
  \label{tab:CDE}
\end{table*}

\subsection{Results and Discussion}
\begin{figure*}[htbp]
    \centering
    \includegraphics[width=0.33\textwidth]{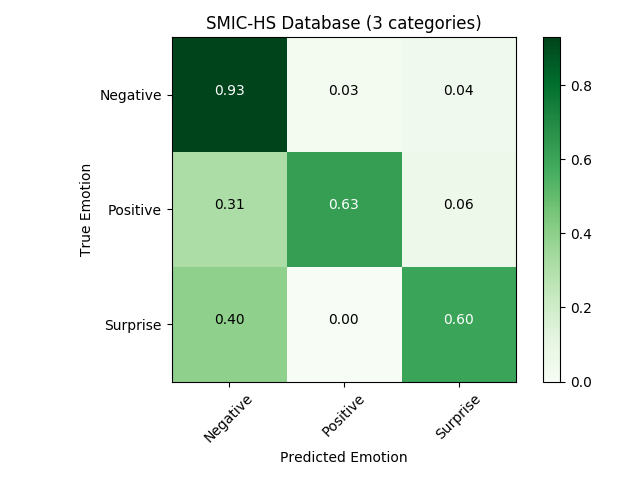}
    \includegraphics[width=0.33\textwidth]{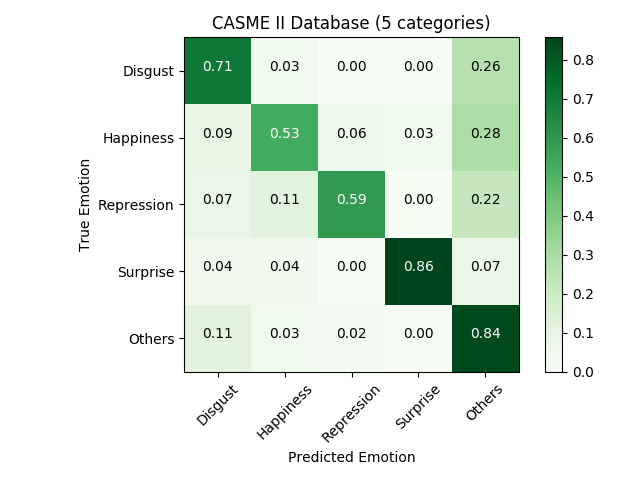}
    \includegraphics[width=0.33\textwidth]{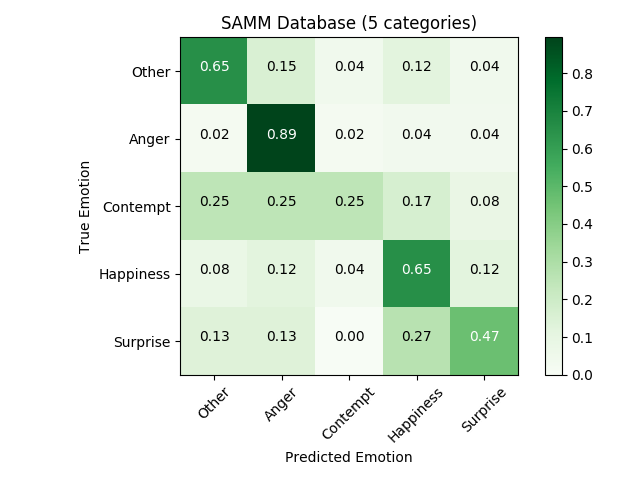}
    \includegraphics[width=0.33\textwidth]{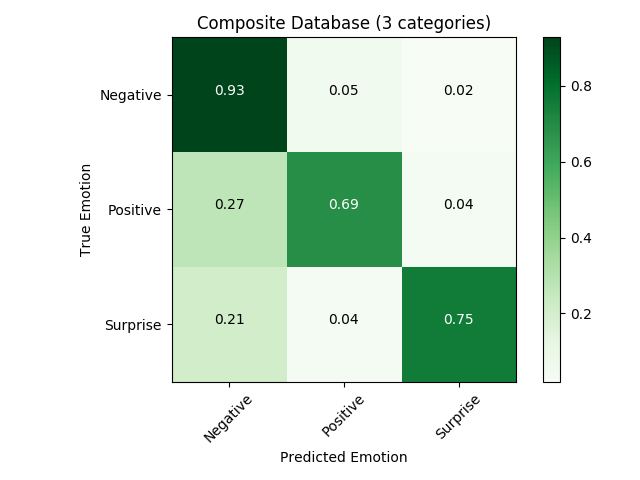}
    \includegraphics[width=0.33\textwidth]{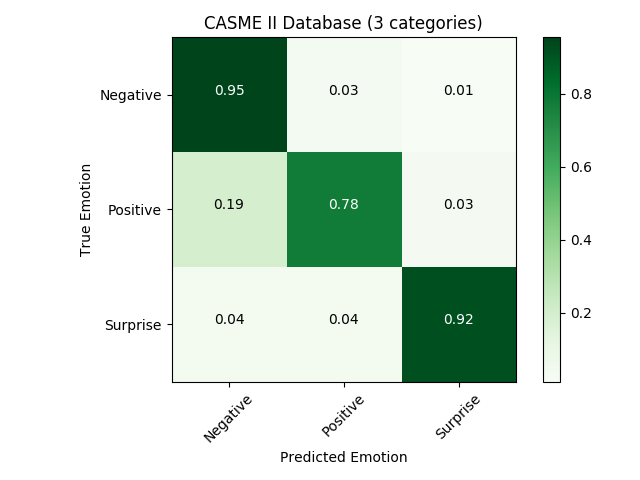}
    \includegraphics[width=0.33\textwidth]{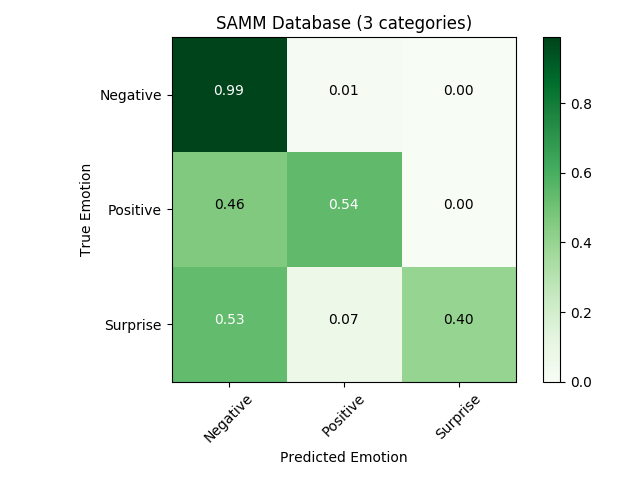}
    \caption{Confusion matrices corresponding to each of our experiments. Only one is shown for SMIC-HS because the SDE and the CDE are identical when this database is used alone.}
    \label{fig:cf_matrix}
\end{figure*}

We compare the performance of the proposed approach with baseline handcrafted feature extraction methods and the most prominent recent deep learning based methods on the widely used micro-expression databases, SMIC-HS, CASME II, and SAMM, described in the previous section, both in the SDE and the CDE settings. To ensure uniformity and fairness of the comparison, the SDE results for all methods were obtained in identical conditions, i.e.\ for the identical number of samples, the number of labels (classes), and using the same cross-validation approach. The details of the performance of our \emph{SLSTT} on different emotion categories are shown in Fig.~\ref{fig:cf_matrix}.

As can be readily seen in Table~\ref{tab:SDE} which presents a comprehensive overview of our experimental results in the SDE setting, the method proposed in the present paper performs best (n.b.\ shown in bold) in all but one testing scenario, in which it is second best (n.b.\ second best performance is denoted by square brackets), trailing marginally behind the method introduced by Sun et al.~\cite{Sun2020}. What is more, in most cases our method outperforms rivals by a significant margin. 

Moving next to the results of our experiments in the CDE setting, these are summarized in Table~\ref{tab:CDE}. It can be readily seen that our method's performance is again shown to be excellent. In particular, in most cases our method again comes out either at the top or second best (as before the former being shown in bold and the latter denoted by square brackets enclosure). The only existing method in the literature which remains competitive against ours is that of Lei et al.~\cite{Lei2021}. To elaborate in further detail, our approach achieved the best results both in terms of UF1 and UAR on CASME II, and on UF1 on the full composite database, and second best on UAR on the composite database and on UF1 on SMIC-HS. The performance of all methods on CASME II is consistently higher than when applied on other data sets, which suggests that the challenge of MER is increased with ethnic diversity of participants -- this should be born in mind in future research and any comparative analysis. It is insightful to observe that in contrast with the results in the SDE setting already discussed (see Table~\ref{tab:SDE}), our method does not come out as dominant in the context of CDE. This suggests an important conclusion, namely that our method is particularly capable of nuanced learning over finer grained classes and that its superiority is less able to come through in a simpler setting when only 3 emotional classes as used. 

Taking into account the results from both the sole and the composite database experiments, it is useful to observe that when only short-range patterns are utilized, convolutional neural network approaches do not outperform methods based on handcrafted feature. It is the inclusion of long-range spatial learning that is key, as shown by the marked improvement in performance of the corresponding methods. Yet, the proposed method's exceeds even their performance, owing to its use of a multi-head self-attention mechanism, thus demonstrating its importance in MER. The superiority of our short- and long-range relation based spatiotemporal transformer is further corroborated by the results shown in the latest two rows in both Table~\ref{tab:SDE} and Table~\ref{tab:CDE} which summarize our comparison of the proposed LSTM aggregator with the simpler mean operator aggregator.

In CASME II, distinguishing whether a micro-expression is Disgust or Others is inherently difficult because the database contains multiple inconsistently labelled samples with only AU4 activated -- some of them are labelled as Others, some as Disgust. It is also worth noting that in SAMM, some AU labels (`AU12 or 14') for the Contempt class were not manually verified, which also causes confusion with the Happiness class (mostly with AU12 labelled). In part, these labelling issues emerge from the fact that the mapping between facial action unit activations and emotions (as understood by psychologists) is not a bijection. It is also the case that imperfect information is made use of because only visual data is used. Hence, it should be understood that the theoretical highest accuracy of automated micro-expression recognition on the MER corpora currently used for research purposes is not 100\%. The micro-expression databases containing multi-modal signals~\cite{Li2022,Li2022a}, which have begun emerging recently, seem promising in overcoming some of the limitations of the existing corpora, and we intend to make use of them in our future work.

\section{Conclusion}
In this paper, we proposed a novel transformer based spatio-temporal deep learning framework for micro-expression recognition, which is the first deep learning work in the field entirely void of convolutional neural network use. In our framework both short- and long-term relations between pixels in spatial and temporal directions of the sample videos can be learned. We use transformer encoder layers with multi-head self-attention mechanism to learn spatial relations from visualized long-term optical flow frames and design a temporal aggregation block for temporal relations. Extensive experimental results using three large MER databases, both in the context of sole database evaluation and composite database evaluation settings and the Leave One Subject Out cross validation protocol, consistently demonstrate that our approach is effective and outperforms the current state of the art. These findings strongly motivate further research on the use of transformer based architectures rather than convolutional neural networks in micro-expression analysis, and we hope that our theoretical contributions will help direct such future efforts.


%



\ifCLASSOPTIONcompsoc
\section*{Acknowledgments}
\else
 \section*{Acknowledgment}
\fi

The authors would like to thank the China Scholarship Council – University of St Andrews Scholarships (No.201908060250) funds L. Zhang for her PhD. This work is funded by the National Key Research and Development Project of China under Grant No. 2019YFB1312000, the National Natural Science Foundation of China under Grant No. 62076195, and the Fundamental Research Funds for the Central Universities under Grant No. AUGA5710011522.

\ifCLASSOPTIONcaptionsoff
 \newpage
\fi



%



\bibliographystyle{IEEEtran}

%








\begin{IEEEbiography}[{\includegraphics[width=1in,height=1.25in,clip,keepaspectratio]{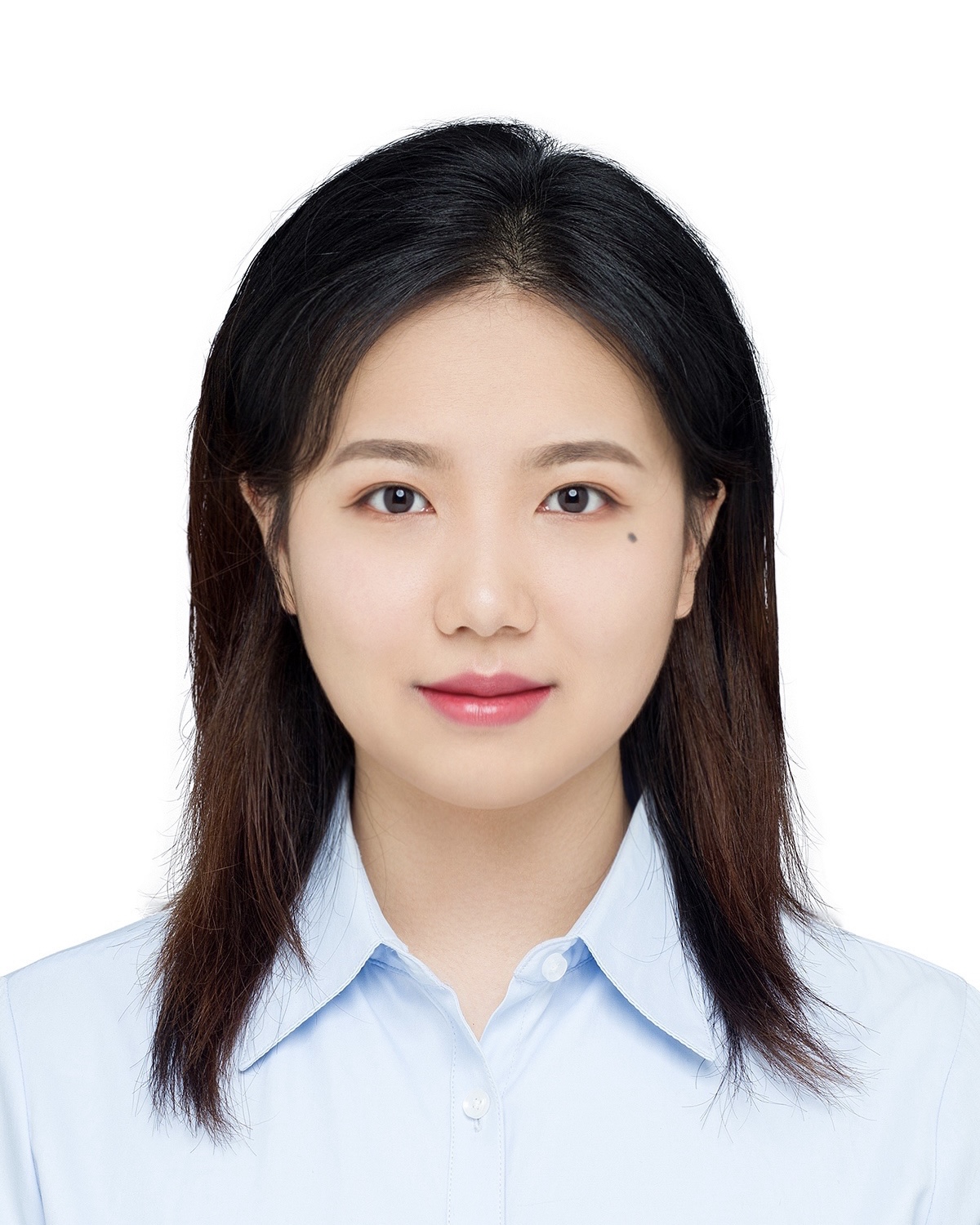}}]{Liangfei Zhang}
received the B.Eng. degree in 2018, from the School of Computer Science and Technology, Northwestern Polytechnical University, P.R.China. She is currently a Ph.D. student in the School of Computer Science, University of St Andrews, where she received her MSc Artificial Intelligence degree in 2019. Her research interests are artificial intelligence, computer vision and pattern recognition, and their application such as affective computing, facial behaviour analysis and expression recognition.
\end{IEEEbiography}

\begin{IEEEbiography}[{\includegraphics[width=1in,height=1.25in,clip,keepaspectratio]{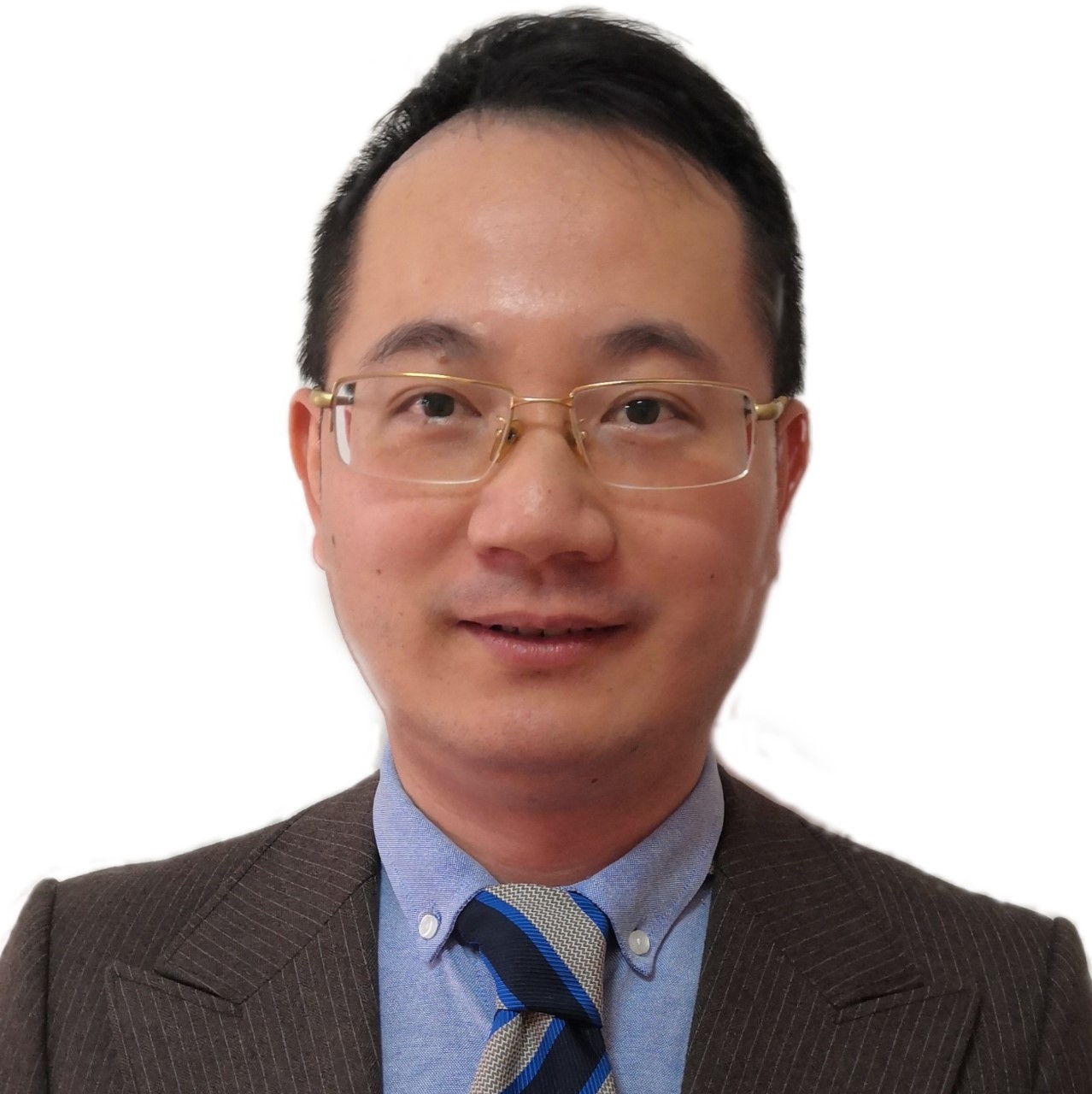}}]{Xiaopeng Hong}
is a professor at Harbin Institute of Technology (HIT), PRC. He had been a professor at Xi'an Jiaotong University, P. R. China, and an adjunct professor with the University of Oulu, Finland. Xiaopeng received his Ph.D. degree in computer application and technology at HIT, in 2010. He has authored over 50 articles in top-tier publications and conferences such as IEEE T-PAMI, CVPR, ICCV, and AAAI. He has served as an area chair/senior program committee member for ACM MM, AAAI, IJCAI, and ICME, a guest editor for peer-reviewed journals like Patter Recognition Letter and Signal, Image and Video Processing, a co-organizer for six international workshops in conjunction with IEEE CVPR, ACM MM, IEEE FG, and a co-lecturer for two tutorials in conjunction with ACM MM21 and IJCB21. His studies about subtle facial movement analysis have been reported by International media like MIT Technology Review and been awarded the 2020 IEEE Finland Section best student conference paper.
\end{IEEEbiography}

\begin{IEEEbiography}[{\includegraphics[width=1in,height=1.25in,clip,keepaspectratio]{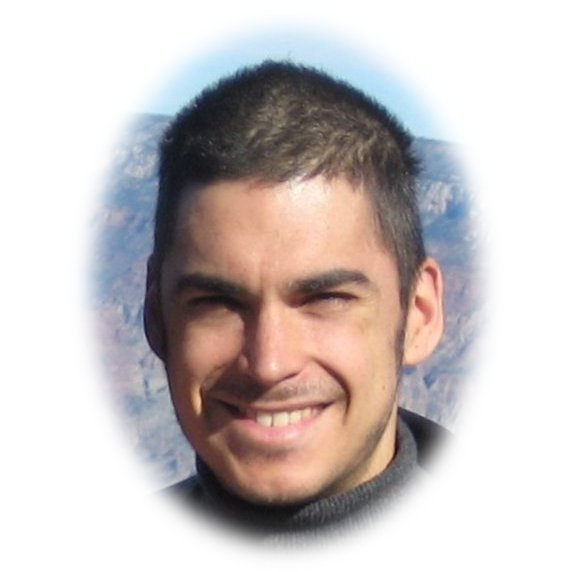}}]{Ognjen Arandjelovi\'c}
graduated top of his class from the Department of Engineering Science at the University of Oxford (M.Eng.). In 2007 he was awarded his Ph.D. by the University of Cambridge where he stayed thereafter as Fellow of Trinity College Cambridge. Currently he is a Reader in the School of Computer Science at the University of St. Andrews in Scotland. Ognjen's main research interests are computer vision and pattern recognition, and their application in various fields of science such as bioinformatics, medicine, physiology, etc. He is a Fellow of the Cambridge Overseas Trust, winner of numerous awards, and Area Editor in Chief of Pattern Recognition and Associate Editor of Information, Cancers, and Frontiers in AI.
\end{IEEEbiography}

\begin{IEEEbiography}[{\includegraphics[width=1in,height=1.25in,clip,keepaspectratio]{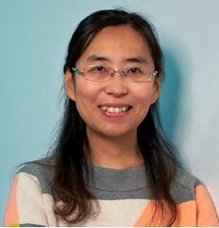}}]{Guoying Zhao}
(IEEE Fellow) is currently an Academy Professor with Academy of Finland and (tenured) full professor (from 2017) with the Center for Machine Vision and Signal Analysis, University of Oulu, Finland. She received the Ph.D. degree in computer science from the Chinese Academy of Sciences and then joined University of Oulu as a senior researcher. She was Academy Fellow in 2011-2017 and an Associate Professor from 2014 to 2017 with University of Oulu. She has authored or co-authored more than 280 papers in journals and conferences. Her papers have currently over 18390 citations in Google Scholar (h-index 65). 
She has served as associate editor for Pattern Recognition, IEEE Transactions on Circuits and Systems for Video Technology, IEEE Trans. on Multimedia and Image and Vision Computing Journals. Her current research interests include image and video descriptors, facial-expression and micro-expression recognition, emotional gesture analysis, affective computing, and biometrics. Her research has been reported by Finnish national TV, newspapers and MIT Technology Review.
\end{IEEEbiography}

\end{document}